# Degendering Resumes for Fair Algorithmic Resume Screening

**Prasanna Parasurama***
New York University
pparasurama@stern.nyu.edu

**João Sedoc**
New York University
jsedoc@stern.nyu.edu

## Abstract

We investigate whether it is feasible to remove gendered information from resumes to mitigate potential bias in algorithmic resume screening. Using a corpus of 709k resumes from IT firms, we first train a series of models to classify the self-reported gender of the applicant, thereby measuring the extent and nature of gendered information encoded in resumes. We then conduct a series of gender obfuscation experiments, where we iteratively remove gendered information from resumes. Finally, we train a resume screening algorithm, and investigate the trade-off between gender obfuscation and screening algorithm performance. We find: (1) There is a significant amount of gendered information in resumes. (2) Lexicon-based gender obfuscation method (i.e. removing tokens that are predictive of gender) can reduce the amount of gendered information to a large extent. However, after a certain point, the performance of the resume screening algorithm starts suffering. (3) General-purpose gender debiasing methods for NLP models such as removing gender subspace from embeddings are not effective in obfuscating gender.

## 1 Introduction

Advances in language models have fundamentally changed the nature of many natural language tasks. In resume screening, for example, simple keyword-based matching has been replaced by more sophisticated NLP models, promising higher quality matches (e.g., Maheshwary and Misra (2018); Lin et al. (2016); Luo et al. (2019)). At the same time, the black-box nature of these models has raised concerns about the potential for bias in downstream applications. For example, in 2018, Amazon came under fire for its resume screening tool that was reportedly biased against women (Dastin, 2018). The model had learned through historical hiring data that men were more likely to be hired, therefore rating male resumes higher than female resumes. Although candidate gender was not explicitly included in the model, it learned to discriminate between male and female resumes based on the gendered information in resumes. For example, men were more likely to use words such as "executed" and "captured".

The bias in this example is due to both the data (i.e. biased hiring data), and the model (i.e. ability of the model to discriminate between genders), both of which are necessary conditions for the overall system bias. A common thread of concern in model-based bias is that more sophisticated models can more easily learn gendered information from resumes, and use the learned gendered information when predicting the outcome of an application (e.g., a resume screening model that takes into account hobbies when predicting the outcome of an application).

In this paper, we address this concern by investigating whether it's feasible to remove gendered information from (i.e., *degender*) resumes to mitigate potential bias in a resume screening algorithm. To do so, we first investigate the extent and nature of gendered information in resumes. Second, we study whether it is possible to obfuscate gender from resumes by conducting a series of experiments, where we iteratively remove gendered information while preserving the job-relevant content. This includes 1. removing gender identifiers such as names, and emails, 2. removing gender indicating words such as "salesman", "waitress", etc. 3. removing hobbies, 4. removing gender-predictive features, 5. removing gender sub-space in embeddings. Finally, we study the trade-off between gender obfuscation and the predictive performance of a resume screening algorithm.

## 2 Related Work

The algorithmic bias literature has proposed several methods to de-bias general-purpose NLP models.

This literature largely focuses on preventing models from inheriting existing societal bias and stereotypes from the training corpus – for example, word embedding models inherit occupational stereotypes and associate "computer programmer" with "man", and "homemaker" with "woman" (Bolukbasi et al., 2016). The de-biasing methods used in this literature can be classified into two broad categories: (1) de-biasing or altering the dataset used to train the model and (2) de-biasing or altering the learning algorithm (See Sun et al. (2019) for a review). Data-based de-biasing methods include, for example, obfuscating or swapping gender in the training data (Zhao et al., 2018a), removing gender subspace in embedding models (Bolukbasi et al., 2016) (See also Gonen and Goldberg (2019) for the shortcomings of this approach), etc. Training-based methods include, for example, putting constraints on the learning algorithm, adversarial learning, etc. (e.g., Zhao et al. (2018b); Zhou and Bansal (2020)). Our proposed de-biasing method in this paper belongs to the first category.

These general-purpose de-biased models address a specific aspect of bias, namely bias reflecting societal bias and stereotypes. This can be helpful in de-biasing resume screening algorithms in some cases. For example, research has shown that gendered wordings exist in job postings (Gaucher et al., 2011; Böhm et al., 2020). If a resume screening algorithm matches resumes to job descriptions using embeddings of the documents based on document similarity, male resumes may be more likely to be matched to job descriptions with masculine language. Here, using a debiased word embedding model may be helpful in de-biasing the screening algorithm. At the same time, there is another type of bias that is relevant in the resume screening context, namely hiring bias in training data, where existing general-purpose de-biased models will likely not be helpful. For example, if the training data contains screening decisions from a gender-biased recruiter, the screening algorithm will learn to use job-irrelevant proxies of gender such as hobbies, style of writing, etc. In this case, de-biasing entails preventing the screening algorithm from using job-irrelevant proxies of gender. Deshpande et al. (2020) use a similar approach in a resume-filtering setting where they down-weight tf-idf scores of features that correlate with nationality.

Finally, a separate set of methods from the algorithmic fairness literature side-step the problem of de-biasing the model completely, but put constraints on the outcomes/decisions of the model to ensure that they are fair based on some fairness criteria (Mitchell et al., 2021). For example, fairness-aware ranking used in LinkedIn Recruiter ensures that the list of candidates returned by a search query has a certain gender distribution (Geyik et al., 2019).

# 3 Data and Methods

The primary dataset is a corpus of applicant resumes from 8 IT firms based in the U.S. These IT firms are clients of an HR analytics firm, which provided us with the aggregated Applicant Tracking System (ATS) data as part of a research partnership. Along with the resume text, we have the applicant's name, gender, years of experience, degree[1], field of study[2], the job posting to which they applied, and the outcome of the application (i.e. whether the applicant received a callback).

## 3.1 Vector Representations of Resumes

In addition to applicant attributes mentioned above, we also extract the skills, competencies, job titles, and job-relevant keywords from the resume using a skills and job titles dictionary[3], and create a dense vector representation of each resume based on these keywords. To do so, we first train a Word2Vec model on resumes to learn a vector representation for all tokens (Mikolov et al., 2013). We then parse the body of the resume into tokens and filter for skills, competencies, job titles, and job-relevant keywords using the dictionary. Finally, we take the average vector representations of the filtered keywords to get one representation for each resume document -– the resulting vector is an embedding of the resume in a skills vector space (See Figure 1).

## 3.2 Matched Sample of Resumes

Occupations vary by gender, so occupational characteristics (i.e. skills, past job experiences, education, etc), are an obvious source of gendered information that a classifier can easily learn. However, such information is less relevant in the context of resume screening applications since applicants applying to the same job are likely to have similar education, skills, and experience. So, to ensure that the classifier learns gendered features beyond occupational characteristics, we match our samples so that male and female resumes are on average similar in observable occupational characteristics. Specifically, we perform 1-1 matching without replacement such that for each male resume, we find a female resume that is within 2 years of experience, has the same degree, field of study, and has a resume similarity score (i.e. cosine similarity of resume vector representations) of at least 0.7. If multiple female resumes match these criteria, we take the resume with the highest similarity score. This matching procedure yields 348k resumes (174k male, 174k female resumes).

---

[1] Associate, Bachelors, Masters, Doctorate
[2] Technical, Science, Business, Law, Other
[3] This dictionary was created by aggregating all the skills and job titles using a secondary LinkedIn dataset.

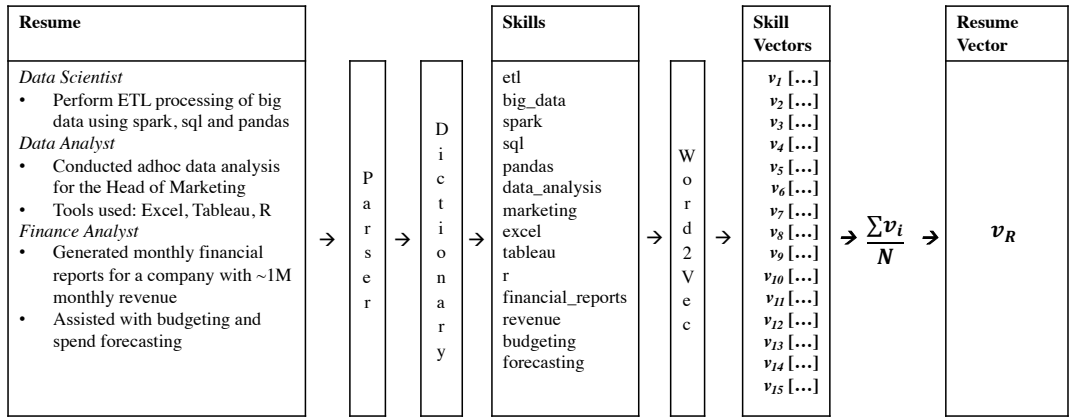

Figure 1: Illustration of Resume Vector Representation

### 3.3 Measuring Gendered Information in Resumes – Gender Classifier

We define gendered information as anything that is predictive of the applicant's gender. To measure the extent of gendered information, we take a predictive modeling approach, where we train a series of models on resumes to classify the gender of the applicant and measure the model's predictive power on a holdout test set.

We use three different sets of models for the classification task: (1) Tf-Idf+Logistic, (2) BigBird, (3) Word Embeddings+Logistic (See Appendix A for model specifications and hyper-parameter tuning). We begin with a bag-of-words baseline using a Tf-Idf+Logistic model. This model is expected to discriminate between genders based on lexical differences. As the main classifier, we use BigBird[4], a state-of-the-art NLP model optimized for long documents (Zaheer et al., 2020). This model is expected to learn to discriminate based on more sophisticated features (e.g., semantic representation, contextual representation, etc.) BigBird uses a variant of the popular BERT-style transformer architecture and is optimized for long documents (Vaswani et al., 2017; Devlin et al., 2019). The classic BERT architecture uses a self-attention mechanism that scales quadratically with the number of tokens in the document. This makes using BERT for long documents such as resumes infeasible due to memory and computational footprint. BigBird overcomes this by using a sparse attention mechanism that scales linearly with the number of tokens in the document. Finally, to compare our results to existing gender de-biased models, we use Word Embedding+Logistic model using both off-the-shelf[5] word embeddings and gender-debiased word embeddings (Bolukbasi et al., 2016). For all of the above models, we use an 80/10/10 train/evaluation/test split.

### 3.4 Gender-Predictive Features – SHAP Values

To understand the gendered information learned by the classification model, we recover the most predictive features using SHAP values (Shapley Additive Explanations) (Lundberg and Lee, 2017). SHAP uses a "perturbation" approach to estimate how a perturbation in the feature space changes the prediction. For example, if we mask a particular token from a resume (e.g. "baseball") and it drastically changes the predicted probability of a given resume, then the token receives a high SHAP value for that particular prediction. In a different resume, removing the same token may not change the predicted probability at all, in which case, the same token receives a SHAP value closer to zero. Although SHAP values provide explainability at the instance level (i.e. for a given token in a given resume), we can aggregate across instances (resumes) to get to model-level feature importance.

---

[4]https://github.com/google-research/bigbird

[5]https://code.google.com/archive/p/word2vec/

### 3.5 Obfuscating Gendered information

Once we understand the extent and nature of gendered information, we investigate whether it's possible to obfuscate gender from resumes while preserving job-relevant content. Keeping in mind the resume screening context, there is a tradeoff between obfuscating gendered information and obfuscating useful job-relevant information. For example, in resume screening, removing all content from resumes except for a handful of job-specific skills and keywords certainly removes gendered information, but at the cost of also removing useful information in the body of the resume. On the other hand, removing names from resumes obfuscates gender without much effect on task-relevant information. First, we remove names (both by string-matching applicant names, and via named entity recognition), emails, LinkedIn IDs, and other URLs from the resume, and replace the tokens with `[DEL]`. Second, we remove gender indicating words such as "salesman", "waitress", etc. (See Appendix B for the full list). Third, we remove hobbies from the resume using the Wikipedia dictionary of hobbies[6]. Fourth, we iteratively remove the most predictive gender features (based on SHAP values) from the resume. Finally, to compare our methods to existing gender de-biased models, we compare gender debiased word embeddings to off-the-shelf word embeddings.

### 3.6 Trade-off Between Gender Obfuscation and Screening Algorithm Performance

As noted earlier, there is a trade-off between gender obfuscation and the performance of the screening algorithm. To understand this trade-off, we train a resume screening algorithm using BigBird, where the input to the model is resume text and job details, and the target is whether the applicant received a callback.

The outcome of an application depends on both the applicant characteristics and the job characteristics – i.e., it depends on the match between

---

the applicant and the job to which they applied. Therefore, the training data should contain both the characteristics of the applicant (resume) and the job (job description). One way to include both the resume characteristics and the job characteristics is to concatenate the resume text and job description text together and feed the concatenated text as a single document to the model. A drawback, however, is that job descriptions tend to be long and full of boilerplate language that does not contain any signal about the outcome of an application. So using the full job description increases the length of each document, which puts unnecessary memory and computational strain on training the model. To overcome this, we get the most important characteristics from the job (company name, job name, business unit, employment type, location, skills, and keywords), concatenate these characteristics with the resume text, and feed the concatenated text as a single document to the model. We get the job characteristics directly from the ATS. For skills and keywords, we use a dictionary of skills and job-relevant keywords (the same dictionary as described in the vector representation section). We then concatenate this text with the resume text to create a single document. The model parses this document into tokens, embeds each token into a vector representation, and creates a tensor representation (an n-dimensional matrix) for the document, which is then fed into a neural network.

## 4 Results

### Gender Obfuscation Experiments

Table 1 reports the out-of-sample gender classification performance using Area Under the Receiver Operating Characteristic (AUROC) scores for a series of obfuscation experiments. AUROC ranges from 0.5 to 1, where 0.5 corresponds to random classification (i.e., no gendered information), and 1 corresponds to perfect classification (i.e., full gendered information). As a more conservative measure, we also calculate the within-job AUROC score, which measures the performance on the subset of applicants that applied to the same job posting. We calculate this score for each job posting separately and aggregate the scores across jobs by taking a weighted average based on the number of applicants in each job.

Gender classification performance decreases as we increasingly remove gendered information. Experiment (1) is the baseline with no matching and no obfuscation, which achieves an AUROC of 0.88. Experiment (2) uses the matched sample, which reduces the AUROC to 0.85. Across (3)-(5), removing names, gender indicating words (IW), and hobbies further reduces AUROC to 0.75. Since experiments (1)-(5) use a bag-of-words Tf-Idf model, the discriminatory features are based on lexical differences between genders. In (6), we replace the Tf-Idf model with a transformer-based BigBird model, which can learn to discriminate on features beyond lexical differences including semantics, style of writing, etc. Indeed, AUROC increases from 0.75 to 0.8.

| | Matched Sample? | Obfuscation | Model | AUROC | With-in Job AUROC |
|---|---|---|---|---|---|
| 1 | No | None | Tf-Idf + Logistic | 0.88 | 0.84 |
| 2 | Yes | None | Tf-Idf + Logistic | 0.85 | 0.83 |
| 3 | Yes | Names/IDs removed | Tf-Idf + Logistic | 0.78 | 0.76 |
| 4 | Yes | Names/IDs, gender IW removed | Tf-Idf + Logistic | 0.75 | 0.73 |
| 5 | Yes | Names/IDs, gender IW, hobbies removed | Tf-Idf + Logistic | 0.75 | 0.72 |
| 6 | Yes | Names/IDs, gender IW, hobbies removed | BigBird | 0.82 | 0.80 |
| 7 | Yes | Names/IDs, gender IW, hobbies, top 100 ftrs removed | BigBird | 0.81 | 0.80 |
| 8 | Yes | Names/IDs, gender IW, hobbies, top 500 ftrs removed | BigBird | 0.79 | 0.77 |
| 9 | Yes | Names/IDs, gender IW, hobbies, top 1k ftrs removed | BigBird | 0.78 | 0.77 |
| 10 | Yes | Names/IDs, gender IW, hobbies, top 2k ftrs removed | BigBird | 0.78 | 0.76 |
| 11 | Yes | Names/IDs, gender IW, hobbies, top 5k ftrs removed | BigBird | 0.76 | 0.74 |
| 12 | Yes | Names/IDs, gender IW, hobbies, top 10k ftrs removed | BigBird | 0.72 | 0.71 |
| 13 | Yes | Names/IDs, gender IW, hobbies, top 20k ftrs removed | BigBird | 0.64 | 0.63 |
| 14 | Yes | Names/IDs, gender IW, hobbies, top 30k ftrs removed | BigBird | 0.60 | 0.57 |
| 15 | Yes | Names/IDs, gender IW, hobbies, top 40k ftrs removed | BigBird | 0.54 | 0.52 |
| 16 | Yes | Names/IDs, gender IW, hobbies removed | Word2Vec + Logistic | 0.68 | 0.65 |
| 17 | Yes | Names/IDs, gender IW, hobbies removed, debiased Word2Vec | Word2Vec + Logistic | 0.67 | 0.64 |

Table 1: Gender Classification Performance on Hold-Out Test Set

To investigate whether further gender obfuscation is possible, we iteratively remove the most predictive gender features (from 100 to 40k) based on Model 6's SHAP values. Experiments (7)-(15) report these obfuscation steps, where, as expected, the performance of gender classification decreases approaching AUROC=0.5 (random classification). Note that in experiments (7)-(15), we do not retrain the model each time. Rather, we remove features in the hold-out test set and evaluate using the model from experiment 6.

Finally, to compare the level of gender obfuscation achieved by existing gender de-biased models, we use gender-debiased word embeddings in experiment (17) (Bolukbasi et al., 2016) and compare it to an off-the-shelf word embedding model on which the de-biased model is based in experiment (16). The de-biased model only reduces the AUROC by a small amount (0.67 vs. 0.68).

### 4.1 Trade-off Between Gender Obfuscation and Screening Performance

To understand the trade-off between gender obfuscation and screening performance, we evaluate the gender-obfuscated hold-out test sets (corresponding to experiments (7) - (15)) using the gender classification model and the screening model. In Figure 2, the left-most point corresponds to the dataset with 100 most predictive gender features removed, and the right-most point corresponds to the dataset with 40,000 most predictive gender features removed.

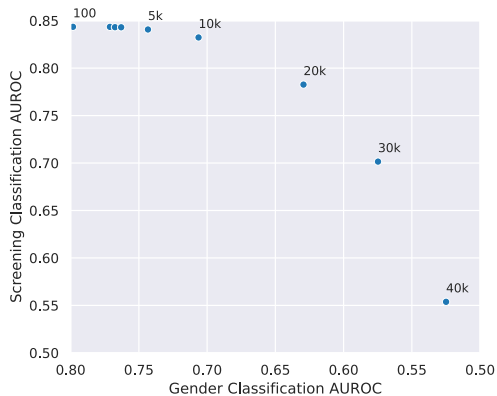

Figure 2: Trade-off Between Gender Obfuscation and Screening Performance

| | All | | Verb | | Adj | | Adv | |
|---|---|---|---|---|---|---|---|---|
| | Male | Female | Male | Female | Male | Female | Male | Female |
| 1 | eagle | hopper | married | greeted | intramural | secretarial | safely | early |
| 2 | lgbt | grace | tasked | piloted | amateur | tumblr | personally | visually |
| 3 | jr | ballet | inspected | volunteered | mechanical | classical | nearly | diligently |
| 4 | scout | actress | said | ordered | myriad | pediatric | routinely | collaboratively |
| 5 | intramural | gamma | prospected | organized | apt | minor | potentially | independently |
| 6 | forklift | wellesley | founded | booked | athletic | elementary | occasionally | functionally |
| 7 | psi | mom | voted | planned | comic | nutritional | virtually | closely |
| 8 | fantasy | receptionist | ranged | brainstormed | numerous | punctual | physically | culturally |
| 9 | chi | pinterest | competed | studied | multivariable | google+ | eventually | promptly |
| 10 | tau | skincare | averaged | experimented | excess | scholastic | collectively | seamlessly |
| 11 | brother | omega | overhauled | exhibited | proud | secret | generally | carefully |
| 12 | dj | lover | positioned | confirmed | recreational | tiny | originally | externally |
| 13 | musician | childhood | lowered | arranged | ll.m | millennial | ultimately | verbally |
| 14 | epsilon | lady | serviced | inspired | solar | flawless | typically | monthly |
| 15 | sergeant | padding | raised | checked | discrete | sick | primarily | smoothly |
| 16 | his | secretarial | shadowed | mailed | armed | administrative | aggressively | extremely |
| 17 | married | summa | saved | scanned | political | facial | additionally | appropriately |
| 18 | songwriter | zeta | surpassed | catered | stony | italian | greatly | accordingly |
| 19 | cars | animals | comprised | learnt | more than 200 | excited | technically | clearly |
| 20 | screenplays | preschool | staged | invited | notable | beautiful | annually | timely |

Table 2: Most Predictive Features for BigBird Gender Classification Model (Model 6)

## 5 Conclusion

There are three important takeaways from these results. First, there is a significant amount of gendered information in resumes. Even after removing names, gender-indicating words and hobbies, a simple Tf-Idf model can learn to differentiate between genders reasonably well (AUC=0.75 in Tf-Idf, AUC=0.82 in BigBird). Second, lexicon-based gender obfuscation method can reduce the amount of gendered information to a certain extent before the performance of the screening algorithm starts suffering. We can reduce the amount of gendered information up to AUC=0.7 without any loss of performance in the screening model. Third, general-purpose gender debiasing methods for NLP models such as removing gender subspace from embeddings are not effective in obfuscating gender.

Within the algorithmic hiring context, these results imply that even a simple resume screening algorithm will learn to differentiate between genders in "anonymized" resumes using gender proxies, and propagate any bias in the training data downstream. Many such algorithms are now being commercialized by HR software firms[7], with little information about how such algorithms deal with potential bias in the training data (Raghavan et al., 2020). Eightfold.ai, a prominent AI talent search software provider, for example, claims that their "AI can help achieve diversity goals by anonymizing people in the hiring process". But, as we have shown, merely taking out names from resumes does not anonymize them from algorithms.

Second, while lexicon-based gender obfuscation method can reduce the amount of gendered information, it is impossible completely hide gender from a screening algorithm without significant costs to the screening performance. In our experiments, experiment (11) achieves perfect gender obfuscation but renders the screening model useless. This calls for active considerations of fairness in algorithmic design such as employing fairness constraints that explicitly take into account gender (Dwork et al., 2012; Geyik et al., 2019; Zehlike et al., 2017). It is important to note that debiasing methods and fairness constraints are not mutually exclusive. One does not render the other moot – it can be desirable to have a screening algorithm that does not use gender-predictive features *and* that has fairness constraints based on outcomes.

---

[7]For example, Eightfold.ai, Findem.ai, Seekout

## A   Model Specifications

For the Tf-Idf + Logistic model, we tried different classifiers including random forest, naive Bayes, SVM, and MLP, and picked the elastic-net logistic regression with mixing parameter `l1=0.5` based on a 5-fold cross-validation.

For the word embedding model, we use Google's Word2Vec model[8] as the baseline, and (Bolukbasi et al., 2016) for the debiased embeddings.

For the BigBird model, we used `Epochs=2`, `N GPUs=2`, `Batch Size per GPU=7`, `Learning Rate=2e-5`, `Weight Decay=2e-5`.

## B   List of Gender Indicating Words

woman man women men womens mens gal guy she he her him girl boy girls boys sorority fraternity female male hostess waitress waiter mother father chairwoman chairman salesman saleswoman

## C   Sample Input Instance for the Screening Model

```
company X
job_loc=
san francisco, ca
job_skills=
build tools, full stack, web development,
impact investing, c, shell, relational
databases, big data, debugging, design,
mobile, python, unix, sql, software
engineering, ruby…
employment_type=
fulltime
source=
jobsite
resume=
john doe
123 center st. new york, ny
education
b.s computer science nyu, ny - may 2015
gpa: 3.6/4
relevant coursework: database design,
operating systems
…
experience
software engineering intern, company y, summer
2013
…
skills
flask, python, keras and ajax
```

Figure 3: Sample Input Data Instance for the Screening Model

---